\documentclass[letterpaper, 10 pt, conference]{ieeeconf}  

\IEEEoverridecommandlockouts                              
                                                          
\usepackage{bm}
\usepackage{cite}
\usepackage{amsmath,amssymb,amsfonts,mathcomp}
\usepackage{algorithmic}
\usepackage{float}
\usepackage{here}
\usepackage{textcomp}
\usepackage[pdftex]{graphicx}

\usepackage{algorithmic}
\usepackage{algorithm}
\usepackage{booktabs}
\usepackage{comment}
\usepackage{url}
\usepackage{hyperref}

\begin{document}

\title{SLAMSpoof: Practical LiDAR Spoofing Attacks on Localization Systems Guided by Scan Matching Vulnerability Analysis
}

\author{Rokuto Nagata$^{1}$, Kenji Koide$^{2}$, Yuki Hayakawa$^{1}$, Ryo Suzuki$^{1}$, Kazuma Ikeda$^{1}$, Ozora Sako$^{1}$, 
\\Qi Alfred Chen$^{3}$, Takami Sato$^{3}$, and Kentaro Yoshioka$^{1}$
\thanks{This research was supported in part by the NSF CNS-2145493, CNS-1929771, USDOT under Grant 69A3552348327 for the CARMEN+ University Transportation Center, JST CREST JPMJCR23M4, JST PRESTO JPMJPR22PA, JSPS KAKENHI 24K02940, and Amano Institute of Technology.}
\thanks{$^{1}$ Department of Electronics and Electrical Engineering, Keio University}
\thanks{$^{2}$ Department of Information Technology
and Human Factors, the National Institute of Advanced Industrial Science and Technology}
\thanks{$^{3}$ Department of Computer Science, University of California, Irvine}
}

\maketitle

\begin{abstract}
Accurate localization is essential for enabling modern full self-driving services. These services heavily rely on map-based traffic information to reduce uncertainties in recognizing lane shapes, traffic light locations, and traffic signs. Achieving this level of reliance on map information requires centimeter-level localization accuracy, which is currently only achievable with LiDAR sensors. However, LiDAR is known to be vulnerable to spoofing attacks that emit malicious lasers against LiDAR to overwrite its measurements. Once localization is compromised, the attack could lead the victim off roads or make them ignore traffic lights. Motivated by these serious safety implications, we design \textit{SLAMSpoof}, the first practical LiDAR spoofing attack on localization systems for self-driving to assess the actual attack significance on autonomous vehicles. SLAMSpoof can effectively find the effective attack location based on our scan matching vulnerability score (SMVS), a point-wise metric representing the potential vulnerability to spoofing attacks. To evaluate the effectiveness of the attack, we conduct real-world experiments on ground vehicles and confirm its high capability in real-world scenarios, inducing position errors of $\geq$4.2 meters (more than typical lane width) for all 3 popular LiDAR-based localization algorithms. We finally discuss the potential countermeasures of this attack. \textbf{Code is available at} \href{https://github.com/Keio-CSG/slamspoof}{\textit{https://github.com/Keio-CSG/slamspoof}}

\end{abstract}


\section{Introduction}
Autonomous driving is one of the most significant technological breakthroughs in the last decades. We can easily take a self-driving taxi service in several cities. The key sensor to enable full self-driving is LiDAR (Light Detection And Ranging) which can obtain the surrounding 3D environment as point clouds. After LiDAR outperformed in the 2007 DARPA Urban Challenge~\cite{urmson2007tartan}, all full self-driving vehicles permitted on public roads in California equip at least one LiDAR sensor on their rooftop~\cite{permitted_ad}.

LiDAR sensing technology is crucial in various aspects of the self-driving pipeline, from obstacle detection to localization systems. Its role is particularly indispensable in localization, which is essential for enabling full self-driving capabilities. Whether using prior map-based methods or Simultaneous Localization and Mapping (SLAM) techniques, these systems heavily rely on LiDAR data to achieve the centimeter-level accuracy required for safe autonomous navigation. This high precision is critical, as the margin between a vehicle and lane markings is typically less than 1 meter~\cite{sato2021dirty}. Such accuracy ensures that self-driving vehicles can operate safely and reliably in complex urban environments.

\begin{figure}[tb]
  \centering
  \includegraphics[trim=0mm 40mm 0mm 15mm,clip,scale=0.30]{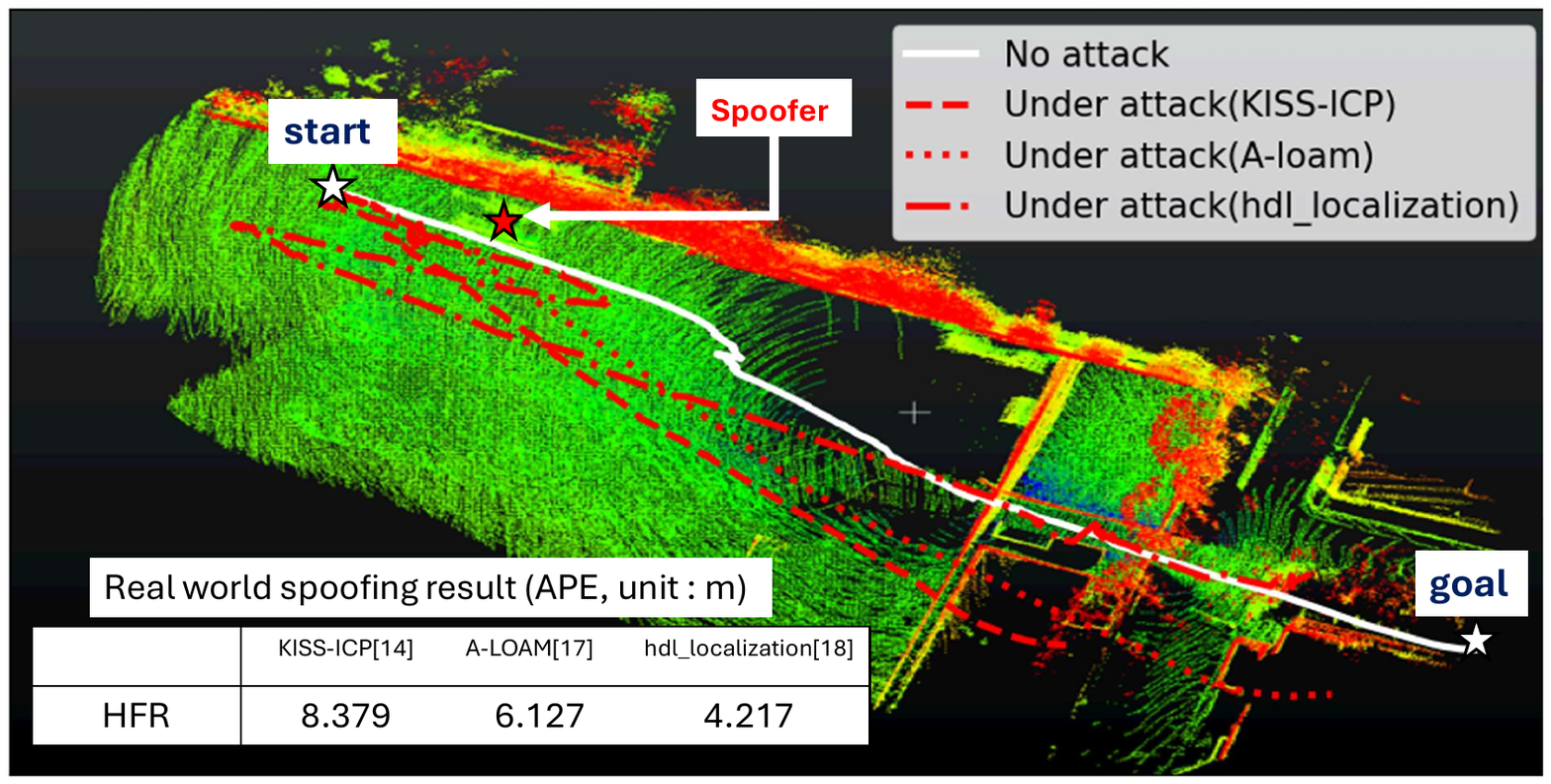}
  \caption{
  \textbf{Real-world attack demo of SLAMspoof on driving vehicle.}
  SLAMSpoof attack successfully deviates the victim vehicle from the planned benign trajectory (while line) to the attack-influenced trajectories (dotted red lines) corresponding to the three major localization algorithms. The induced position errors are $\geq$4.2 meter, which is wider than the typical lane width as shown in the lower-left table.
  }
  \label{header_fig}
\end{figure}


Despite the critical role of LiDAR in self-driving, recent lines of research have reported the vulnerabilities of LiDAR against spoofing attacks~\cite{petit2015remote, cao2019adversarial, sato2024lidar}, which shoot malicious lasers with more power than legitimate ones to overwrite the measurements. With LiDAR spoofing attacks, the adversary can inject ghost points into the point cloud or remove a part of the point cloud.
Specifically, prior work~\cite{sato2024lidar} demonstrates that removal attack is effective on state-of-the-art LiDARs, removing almost all points in 80$^\circ$ horizontal range.

Nevertheless, none of the prior works have successfully demonstrated LiDAR spoofing attacks on LiDAR-based localization for driving autonomous vehicles~\cite{xu2023sok}. Yoshida et al,~\cite{yoshida2022adversarial} shows that LiDAR-based SLAM could be vulnerable to malicious manipulation on the point cloud, but this work targets only 2D-LiDAR and does not mention how to achieve such malicious manipulation. Fukunaga et al.~\cite{fukunagarandom} demonstrate that random LiDAR spoofing attacks can compromise the point cloud environmental map, but this work does not target the online localization for the driving vehicle. Furthermore, this work admits that their random attack cannot cause major attack impacts on the x (longitudinal) and y (lateral) axes.

Motivated by this, we conduct the first security analysis of LiDAR-based localization systems against LiDAR spoofing attack for driving autonomous vehicles, and design \textit{SLAMSpoof}, the first practical LiDAR spoofing attacks on LiDAR-based localization for self-driving to assess the actual attack significance on autonomous vehicles.
Our study has the following contributions:
\begin{itemize}
    \item We introduce \textit{Scan Matching Vulnerability Score (SMVS)}, a metric to quantify the scan matching's vulnerability to spoofing, establishing a new standard for security evaluation in localization systems.
    \item We develop SLAMspoof, a framework centered on SMVS, which significantly enhances real-world attack feasibility compared to conventional methods.
    \item We successfully demonstrate the first real-world spoofing attack on the self-positioning system of a LiDAR-equipped robot during long-distance travel.
    \item We discuss potential defenses against SLAMspoof.
\end{itemize}

\section{Related work and Background}
\subsection{LiDAR Spoofing Attacks} \label{sec:lidar_spoofing}
\begin{figure}[tb]
  \centering
  \includegraphics[trim=25mm 40mm 0mm 30mm,scale=0.36]{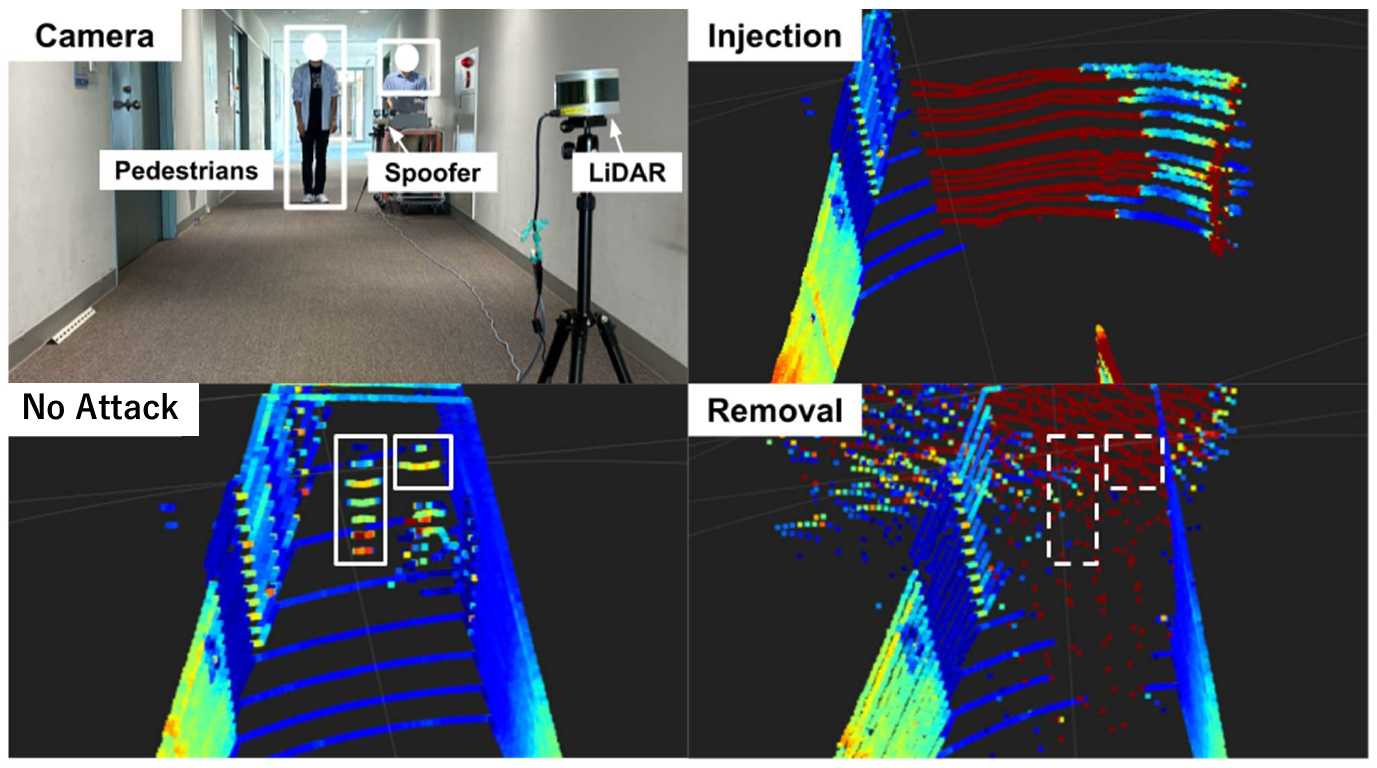}
  \caption{The effects of spoofing attacks on LiDAR. In an injection attack (top-right), false point cloud data representing a non-existent wall is inserted into the LiDAR scan. In a removal attack (bottom-right), injected noise obscures the pedestrians' point cloud, effectively erasing them from the LiDAR scan.}
  \label{example_spoofing}
\end{figure}

LiDAR is fundamentally vulnerable to other malicious laser sources as the nature of the LiDAR mechanism that senses the laser reflection. This type of attacks is known as LiDAR spoofing attacks~\cite{shin2017illusion, petit2015remote, cao2019adversarial}, where Fig.\ref{example_spoofing} shows the point cloud under attack. Based on the attack effects, there are two major types of LiDAR spoofing attacks: injection and removal attacks. For injection attacks, once initial works~\cite{shin2017illusion, cao2019adversarial} demonstrated that their attacks can inject a few hundred points on the point cloud, the state-of-the-art attack~\cite{sato2024lidar} can compromise 99\% of points within 83$^\circ$ ($\sim$7k points). 
On the other hand, removal attacks~\cite{shin2017illusion, cao2023you,sato2024lidar} try to erase the legitimate points from the detected point clouds. The state-of-the-art removal attack, HFR attack~\cite{sato2024lidar}, can remove almost all points within 80$^\circ$ range. 

\subsection{LiDAR-based Localization} \label{sec:lidar_slam}
Localization is a methodology to identify the ego location on a map with sensor inputs. While the multi-sensor fusion (e.g., LiDAR-IMU fusion) approach has been gaining popularity owing to its accuracy and robustness~\cite{fastlio2,koide2024_02}, the localization solely with LiDAR is still popular due to its simplicity and applicability~\cite{vizzo2023ral, cticp, mulls}. In this study, we focus on the three popular LiDAR-based localization algorithms, LiDAR-based localization methods, namely A-LOAM~\cite{zhang2014loam}, KISS-ICP~\cite{vizzo2023ral}, and hdl\_localization~\cite{koide2019portable}, covering three major point distance calculations (plane and edge feature matching~\cite{zhang2014loam}, ICP~\cite{zhang1994iterative}, NDT~\cite{magnusson2009three}) and two major scan matching schemes (online local map and using prior-map).
This study is the first to explore the vulnerability of these major LiDAR-based localization methods against LiDAR spoofing attacks in driving vehicles.

\subsection{Prior Attack Attempts on LiDAR-based Localization} \label{sec:prior_attack}
As also discussed in~\cite{xu2023sok}, no successful LiDAR spoofing attack was demonstrated on LiDAR-based SLAM for driving autonomous vehicles. Yoshida et al.~\cite{yoshida2022adversarial} showed that LiDAR-based SLAM could be vulnerable to a malicious point injection. However, this work only evaluates their attack on 2D-LiDAR, which cannot support autonomous driving, in a lab-level scenario. 
Fukunaga et al.~\cite{fukunagarandom} demonstrate random LiDAR spoofing attacks to compromise the prior environment map generation. This work does not target to attack the online localization systems in the driving vehicle and admits that their random attack cannot cause major attack impacts on the x (longitudinal) and y (lateral) axes. 


\subsection{Threat Model and Attack Goal}
We assume that the attacker can place a LiDAR spoofing device on a roadside where the victim robot will pass through, which is known as ``Spoofer placed in environment'' threat model~\cite{hallyburton2022security}. We assume that the victim uses a 3D LiDAR such as Velodyne VLP-16~\cite{VLP16} and Livox Horizon~\cite{livox_horizon} and solely relies on LiDAR-based localization without sensor fusion with other sensors such as IMU and GNSS.  
We assume robots that follow a fixed route (e.g., autonomous bus) as the attack target, and the attacker can know the route where the victim will pass by.
The attacker can know which LiDAR is used in the victim vehicle based on the appearance or some documents. However, we assume that the attacker cannot know which localization algorithm is used by the victim. The attacker aims to deviate the victim from their driving lane. According to~\cite{sato2021dirty}, the deviation needs 0.29 meters on local roads and 0.74 meters on highways to touch the lane line.


\section{Methodology: SLAMSpoof Attack}

\begin{figure*}[h]
  \includegraphics[width=0.85\linewidth]{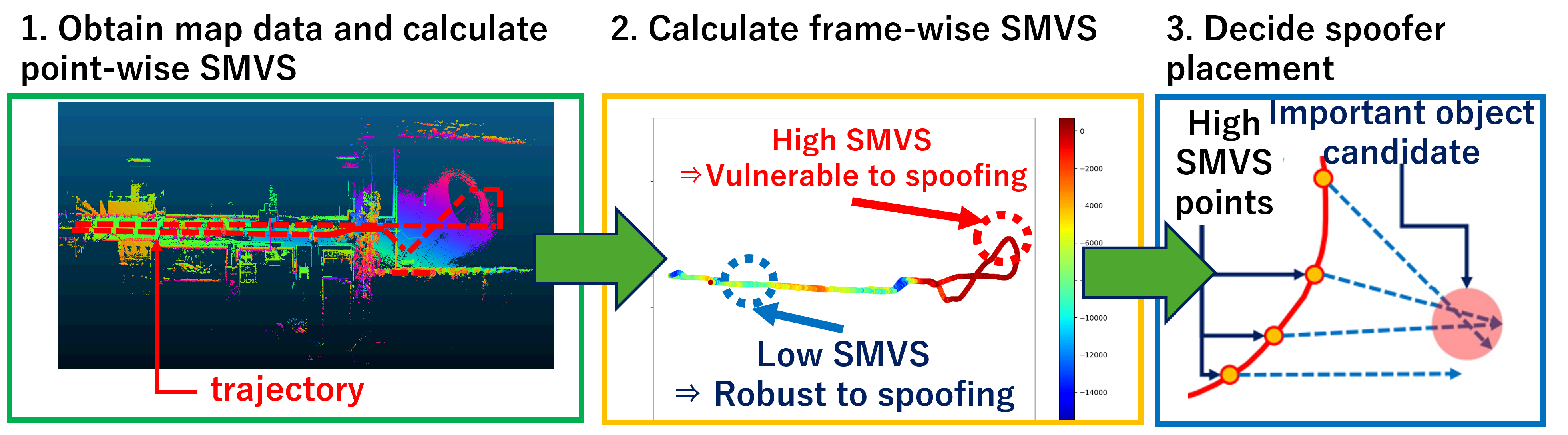}
  \centering
  \vspace{-0.1in}
  \caption{The overview of the \textit{SLAMSpoof} framework, which is based on Scan Matching Vulnerability Score (SMVS). First, the attacker replicates the target's route to acquire map data. The SMVS distribution is generated from the map data to identify the optimal attack location. The calculation methods for point-wise SMVS are described in III-B-1, for frame-wise SMVS in  \S \ref{sec:Frame-wise}, and the process of determining the spoofer placement is detailed in \S \ref{sec:attack_loc_sel}.}
  \label{proposed_method_concept}
  \vspace{-0.1in}
\end{figure*}

To systematically evaluate the security of LiDAR-based localization in driving vehicles, we design the SLAMSpoof attack. As discussed in~\S\ref{sec:prior_attack}, prior attack attempts with random spoofing attacks were not successful due to the robustness of LiDAR-based localization. This result motivates us to design an effective but lightweight methodology to prioritize the potential attack locations based on their vulnerability level because reckless attack attempts are likely to be a waste of time and effort.. Fig.~\ref{proposed_method_concept} illustrate the overview of SLAMSpoof consisting of 3 steps: the attacker \textbf{(1)} first collects point cloud data of the entire potential routes where the victim will likely drive and calculate point-wise SMVS representating of the point importance on scan matching, \textbf{(2)} calculates the frame-wise SMVS, which quantifies the potential vulnerability with the point cloud of the frame, and \textbf{(3)} finally finds the best location to launch LiDAR spoofing attack. We will explain each step in the rest of this section.


\subsection{Point-Wise SMVS Calculation}

Localization systems are designed to be robust against partial occlusions, making random point cloud alterations ineffective. Efficient attacks require identifying and manipulating critical areas that play key roles in scan matching. Scan matching determines pose by estimating optimal translation and rotation parameters between two point clouds, controlled by geometric constraints. We introduce point-wise SMVS, a metric that identifies feature points significantly affecting geometric consistency. It quantifies each point's influence on scan matching to calculate its importance.

For the point-wise SMVS, analyzing the Hessian matrix of the scan matching objective function is crucial to identify vulnerabilities. The eigenvector corresponding to the minimum eigenvalue of this Hessian indicates the weakest constraint direction in pose estimation. Conversely, the eigenvector of each point's Hessian's maximum eigenvalue shows the direction of its strongest constraint.
Combining these insights quantifies each point's contribution to overall scan matching, revealing vulnerable directions and points strongly influencing constraints in those directions. Strategic manipulation of these critical points to weaken constraints can maximize attack effectiveness on scan matching.
To analyze point-wise SMVS for a frame, two point clouds are generated by applying noise and random sampling. These are then treated as target and source point clouds for analysis.

Consider a point cloud $\mathcal{P} = \{ \mathbf{p}_i \in \mathbb{R}^3 |_{i=1, 2, \cdots N} \}$. Pose estimation is performed by estimating the pose $\mathbf{T}$ that minimizes the distance residual $e(\mathbf{T} \mathbf{p}_i)$ between the source point $\mathbf{p}_i$ and its nearest target point. In general, the minimization of the residual is achieved through iterative optimization. The update $\delta \mathbf{T}$ to the pose $\mathbf{T}$ is obtained using the Gauss-Newton method as follows.
\begin{align}
  \mathbf{J}_i &= \frac{\partial}{\partial {\mathbf T}} e(\mathbf{T} {\mathbf p}_i), \\
  \mathbf{H}_{\text{global}} &= \sum {^i\mathbf{H}_\text{local}} = \sum \mathbf{J}_i^T \mathbf{J}_i, \\
  \mathbf{b}_{\text{global}} &= \sum \mathbf{J}_i^T e(\mathbf{T} {\mathbf p}_i), \\
  \delta \mathbf{T} &= \mathbf{H}_{\text{global}}^{-1} {\mathbf b}_{\text{global}}.
\end{align}

Here, when $\mathbf{H}_{\text{global}}$ is not invertible, i.e., when the smallest eigenvalue is sufficiently small, the update vector $\delta \mathbf{T}$ diverges, making the pose estimation unstable. By finding the smallest eigenvalue and its corresponding eigenvector from the point cloud, and determining the strength of the constraint in that direction, the influence of each point on the update of the estimated pose can be quantified. Although the specific objective function varies depending on the scan matching method, it is assumed that most of them show a similar trend in the influence of each point since they are based on the matching of local geometric shapes. Here, we use the representative distribution-to-distribution matching function of Generalized ICP \cite{segal2009generalized}.

Let the eigenvalues and eigenvectors of the Hessian matrix $\mathbf{H}_\text{global}$ of a scan matching result be $\lambda^{j}_{\text{global}}, \mathbf{x}^{j}_{\text{global}}$. Also, let $\mathbf{x}^{\text{min}}_{\text{global}}$ be the eigenvector corresponding to the smallest eigenvalue. Let the Hessian matrix of point $\mathbf{p}_{i}$ be $^{i}\mathbf{H}_{\text{local}}$, and its eigenvalues and eigenvectors be $\lambda_{\mathbf{p}_i}^j, \mathbf{x}_{\mathbf{p}_i}^j$. The eigenvector $\mathbf{x}^{\text{max}}_{\mathbf{p}_{i}}$ corresponding to the largest eigenvalue represents the direction in which the point is most strongly constrained. By combining this information, the importance of point $\mathbf{p}_i$ is obtained as $I_{\mathbf{p}_i} = |\mathbf{x}^{\text{max}}_{\text{global}} \cdot \mathbf{x}^{\text{min}}_{\mathbf{p}_{i}}|$. The procedure for calculating the point-wise SMVS is summarized in Algorithm \ref{alg1}.


\begin{algorithm}[tb]
    \caption{Calculation of point-wise SMVS}
    \label{alg1}
    \begin{algorithmic}[1] 
    \STATE $\mathbf{H}_{\text{global}} \leftarrow  \text{Hessian matrix of a scan matching result}$
    \STATE $\mathbf{x}^{\text{min}}_{\text{global}}, {\lambda}^{\text{min}}_{\text{global}} \leftarrow \text{Minimum eigenvector/value of } \mathbf{H}_{\text{global}}$
    \FOR{$\mathbf{p}_i \in \mathcal{P}$}
    \STATE $^{i}\mathbf{H}_{\text{local}} \leftarrow  \text{Local Hessian matrix of} \, \mathbf{p}_i$
    \STATE $ \mathbf{x}^{\text{max}}_{\mathbf{p}_i}, {\lambda}^{\text{max}}_{\mathbf{p}_i} \leftarrow \text{Maximum eigenvector/value of } {^{i}\mathbf{H}_{\text{local}}}$
    \STATE $ I_{\mathbf{p}_i} \leftarrow | \mathbf{x}^{\text{min}}_{\text{global}} \cdot \mathbf{x}^{\text{max}}_{\mathbf{p}_i}|$ 
    \ENDFOR
    \end{algorithmic}
\end{algorithm}

\subsection{Frame-Wise SMVS Calculation}\label{sec:Frame-wise}

\begin{figure}[tb]
  \centering
  \includegraphics[width=0.9\linewidth]{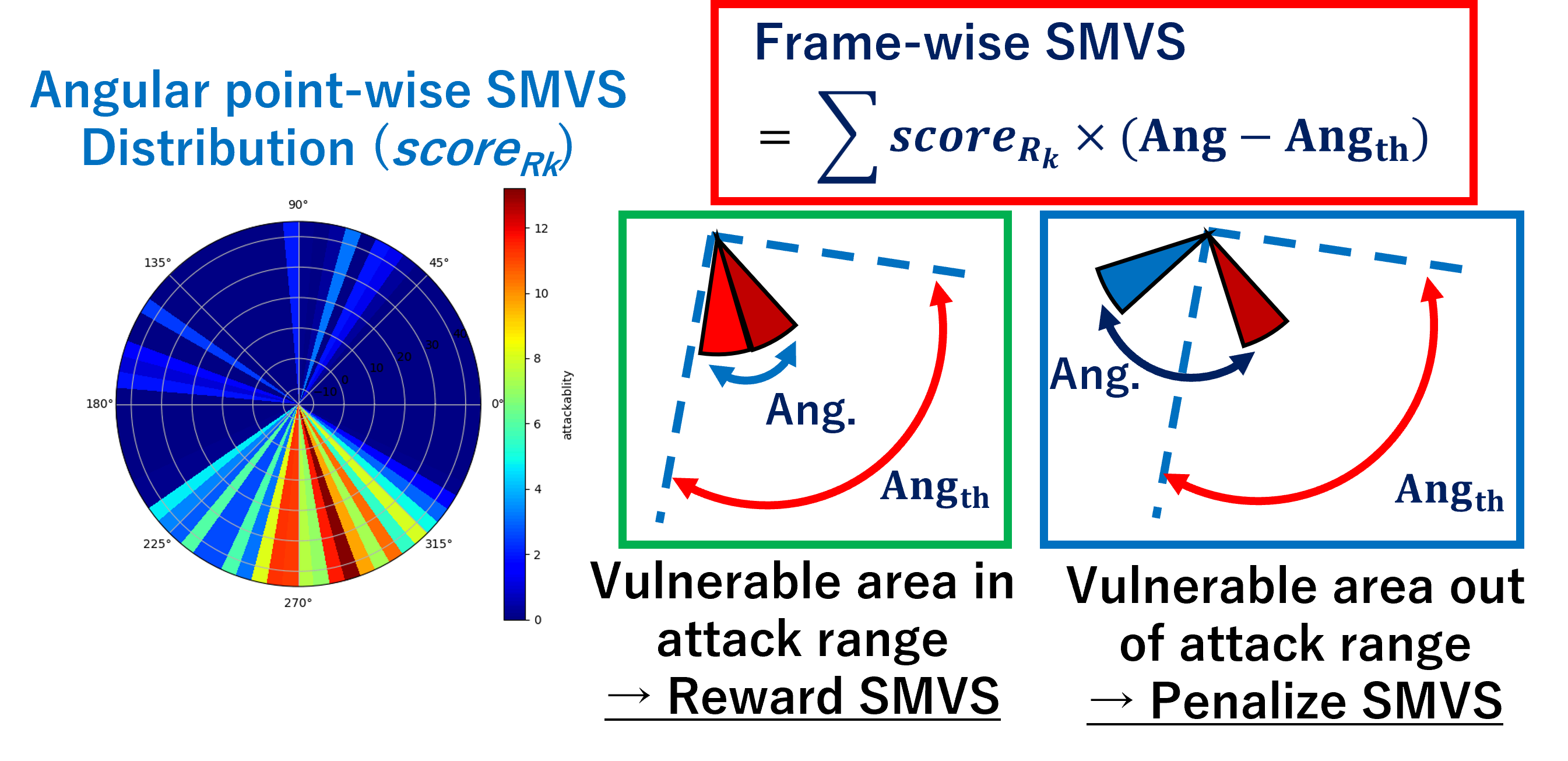}
  \vspace{-0.1in}
  \caption{Frame-wise SMVS calculation. We create an angular polar plot from point-wise SMVS ($score_{Rk}$), and then the frame-wise SMVS is calculated if the region is within the attack range or not, for each angular region.}
  \label{}
\end{figure}

In real-world LiDAR spoofing attacks, deploying multiple attack devices is impractical, limiting horizontal tampering range. This makes attacking localization systems difficult when high-importance points are widely dispersed (e.g., robots moving between buildings). To quantify this vulnerability, we propose the \textit{frame-wise} SMVS metric.
Frame-wise SMVS is designed to yield high scores when critical point clusters are concentrated in one area. Specifically, it outputs high values when a large number of point-wise SMVS-weighted points are concentrated within the spoofing range, and low values when dispersed.
By continuously evaluating frame-wise SMVS along a trajectory, we can identify the most effective attack points for a single spoofer. This approach accounts for real-world constraints and enables strategic targeting of localization systems.

To calculate frame-wise SMVS, the point cloud is segmented based on the horizontal azimuth angle of each point. The horizontal azimuth angle $\theta_{\mathbf{p}_i}$ of point $\mathbf{p}_i$ is
\begin{align}
\theta{\mathbf{p}_i} = \arctan{\frac{y_{\mathbf{p}_i}}{x_{\mathbf{p}_i}}},
\end{align}
where $x_{\mathbf{p}_i}$ and $y_{\mathbf{p}_i}$ are the $x$- and $y$-coordinates of point $\mathbf{p}_i$, respectively. The point cloud is divided into $n$ small regions, and each region is denoted as $R_{k}$. Each region contains points $\tilde{\mathcal{P}}_k$ whose horizontal angles are in the range $\frac{2 \pi k }{n}$ to $\frac{2 \pi (k + 1)}{n}$ radians, where $k = 0, \cdots, n - 1$. Then, the score of $R_k$ is computed as
\begin{align}
  \mathrm{score}_{R_k} = \sum_{\mathbf{p}_m \in \tilde{\mathcal{P}}_k} I_{\mathbf{p}_m}.
\end{align}
where $I_{\mathbf{p}_m}$ represents the point-wise SMVS.

To evaluate vulnerability with physical spoofing constraints, we calculate angular differences from a central spoofing angle. This central angle corresponds to the region with the highest score. We define the distance $d_k$ between a region $R_k$ and the center as:
$$d_k = |k_{\text{center}}-k| \pmod n$$
where $k_{\text{center}}$ is the index of the region with the highest score.

Frame-wise SMVS is determined by combining each region's score with its attack feasibility. The feasibility primarily depends on the horizontal angular constraint, with larger $d_k$ indicating increased attack difficulty. We designed a distance function $f(d_k)$ to reflect this relationship:
$$f(d_k) = -d_k + d_{\text{th}}$$
where $d_{\text{th}}$ is a threshold distance.
Consequently, we define frame-wise SMVS as:
$$ \text{Frame-wise SMVS} = \sum_{k=0}^{n-1} \mathrm{score}_{R_k} \cdot f(d_k)$$

\begin{figure}[tb]
  \includegraphics[trim=0mm 40mm 0mm 50mm, clip,scale=0.3]{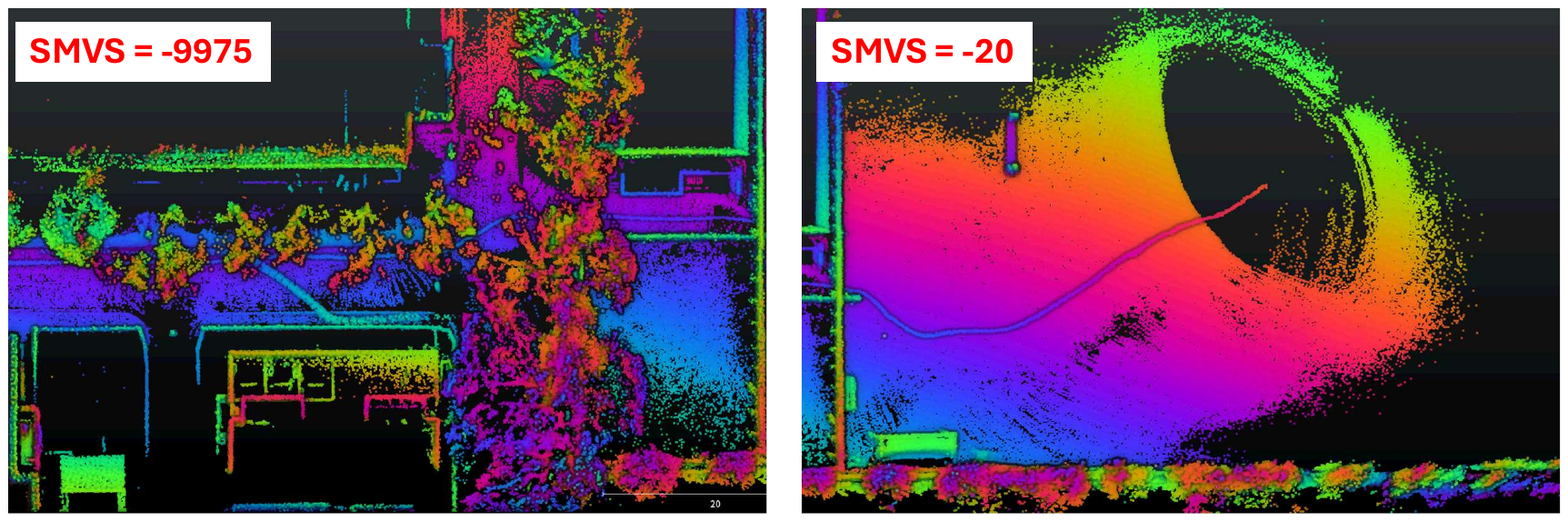}
  \vspace{-0.5in}
  \caption{(Left) Example of low SMVS. The point cloud is distributed across a wide range of directions, allowing accurate pose estimation from the point cloud outside the spoofing range. (Right) Example of high SMVS. The directional distribution of the point cloud is biased, making it vulnerable as almost all points can be tampered with by spoofing. }
  \label{example_smvs}
\end{figure}

In scenes where features are distributed across a wide range of horizontal angles, such as areas surrounded by buildings (Fig. \ref{example_smvs}, left), the frame-wise SMVS becomes small. In these cases, critical data for scan matching is dispersed, allowing correct matching from unaltered points outside the attack range, even if a portion is tampered with.
Conversely, in open areas (Fig. \ref{example_smvs}, right), important matching points tend to cluster, resulting in a higher frame-wise SMVS.
This frame-wise SMVS metric can be utilized to preemptively identify vulnerable regions along a robot's trajectory that are susceptible to spoofing attacks.

\subsection{Attack Location Selection}\label{sec:attack_loc_sel}


Finally, we propose a method to determine the effective spoofer placement by leveraging the frame-wise SMVS analysis. We first identify frames with high frame-wise SMVS scores and estimate the direction of critical point clusters that strongly influence scan matching in these frames. By drawing half-lines in these estimated directions and calculating their intersection points across all frames, we can pinpoint locations of objects crucial for scan matching. These intersections indicate optimal positions for spoofer placement, as tampering with these critical objects is likely to maximize the impact of the attack on position estimation.

To determine spoofer placement from line intersections, we first remove outliers from intersection coordinates, considering points within $\pm2\sigma$ of the mean as potential critical object candidates. After outlier removal, we calculate the intersection points' bounding box and approximate the trajectory with a linear function.
We derive this approximation using least squares method on the top SMVS points, resulting in a function $f(x) = mx + n$. The perpendicular line passing through the bounding box center $(C_x, C_y)$ is represented as: 
\begin{align}
g(x)=-\frac{1}{m}x+(-mC_x + C_y),
\end{align}
This line serves as the candidate for spoofer placement. We use the bounding box center to ensure a wide central area is tampered with, allowing flexibility in placement due to real-world constraints. In practical applications, the spoofer should be placed 10-15m away from the trajectory intersection point, balancing spoofing duration and effectiveness.

\section{Evaluation}

We first evaluate whether our SMVS can effectively prioritize locations based on their vulnerability level, and then evaluate 
the actual attack effectiveness of the SLAMSpoof on the three major LiDAR-based localization methods in autonomous driving scenarios not only with a simulator but also with physical-world experiments.


\subsection{SMVS Validity Evaluation}
\vspace{-0.09in}

\noindent \textbf{Experimental Settings:}
To verify if our SMVS can effectively prioritize attack locations based on their vulnerability, we simulated 16 different attack device placement patterns along the course similar to Fig.~\ref{header_fig}, each designed to yield various SMVS. For each placement position, we ran 10 simulations to measure the trajectory deviation caused by LiDAR spoofing. Here, we evaluate via RMSE of the Absolute Pose Error (APE)~\cite{zhang2018tutorial} and the maximum value of Relative Pose Error (RPE)~\cite{zhang2018tutorial}.


In our experiments, we modeled three patterns of LiDAR spoofing attacks based on the results in~\cite{sato2024lidar}.
The High-Frequency Removal (HFR) Attack, which uses high-frequency pulses to alter existing point clouds, replaces points within an 80$\tcdegree$ range with salt-and-pepper noise. Since such noise is easily removed by preprocessing pipelines, we also simulated a version where points within this angular range were simply removed.
For the injection attack, we simulated the appearance of a wall at a fixed distance, spanning 80$\tcdegree$ degrees, as in Fig.\ref{example_spoofing}.

To verify the generality of our proposed method, we conducted simulations on three major localization techniques: KISS-ICP~\cite{vizzo2023ral}, A-LOAM~\cite{zhang2014loam}, hdl-localization~\cite{koide2019portable}. Since hdl-localization requires a prior environment map, we first ran A-LOAM without any attacks to generate the necessary map data, which was then used during the execution.

We assumed that localization methods considering local plane shapes behave similarly. The importance of each point was calculated based on point cloud matching using GICP~\cite{segal2009generalized}. To calculate SMVS, we divided the point cloud into 72 small regions, thus each $R_k$ has an angular range of 5\textdegree.
We model that the removal attack's range is approximately $\pm40\tcdegree$ from the central angle, corresponding to 8 adjacent regions from the center. Consequently, we set $d_{th} = 8$ to impose penalties outside this attack range.

\begin{table*}[tb]
\centering
\caption{SMVS-APE relation. $\mathcal{S}$ means SMVS value.}
\vspace{-0.1in}
\label{tab_APE}
\begin{tabular}{c|cccccc}
\toprule
\multicolumn{7}{c}{KISS-ICP\cite{vizzo2023ral}} \\ \midrule
SMVS                                             & HFR (no noise) [m]    & HFR (noise) [m]       & Injection [m]        & HFR (no noise) [deg]  & HFR (noise) [deg]     & Injection [deg]      \\ \midrule
-10000 \textless $\mathcal{S}$ \textless -6000 & 0.164±0.057          & 0.191±0.065          & 0.623±0.182          & 0.755±0.815          & 0.905±0.771          & 1.801±0.617          \\
-6000 \textless $\mathcal{S}$ \textless -3000  & 0.179±0.075          & 0.212±0.135          & 0.946±0.069          & 1.223±1.743          & 1.495±1.547          & 4.609±1.945          \\
-3000 \textless $\mathcal{S}$ \textless -1000  & 0.167±0.057          & 0.264±0.074          & 7.732±0.947          & 0.911±0.964          & 1.996±1.321          & 13.04±3.087          \\
-1000 \textless $\mathcal{S}$                 & \textbf{1.323±1.492} & \textbf{8.800±3.963} & \textbf{16.74±2.071} & \textbf{4.191±3.642} & \textbf{39.28±38.59} & \textbf{30.80±10.57} \\ \midrule \addlinespace \midrule
\multicolumn{7}{c}{A-LOAM\cite{zhang2014loam}} \\ \midrule
SMVS                                             & HFR( no noise ) [m]    & HFR (noise)  [m]       & injection [m]        & HFR( no noise ) [deg]  & HFR (noise)  [deg]     & injection [deg]      \\ \midrule
-10000 \textless $\mathcal{S}$ \textless -6000 & 0.155±0.095          & 0.152±0.058          & 0.124±0.069          & 1.206±0.552          & 1.247±0.779          & 1.021±0.631          \\
-6000 \textless $\mathcal{S}$ \textless -3000  & 0.130±0.037          & 0.154±0.026          & 0.089±0.014          & 1.525±0.321          & 3.635±0.997          & 1.229±0.050          \\
-3000 \textless $\mathcal{S}$ \textless -1000  & 0.344±0.072          & 0.319±0.083          & 0.357±0.080          & 4.830±0.775          & 6.542±2.507          & 4.551±0.666          \\
-1000 \textless $\mathcal{S}$                  & \textbf{3.135±4.545} & \textbf{2.608±3.206} & \textbf{4.094±3.595} & \textbf{55.08±53.45} & \textbf{53.86±48.93} & \textbf{48.91±41.97} \\ \midrule \addlinespace \midrule
\multicolumn{7}{c}{hdl\_localization\cite{koide2019portable} } \\ \midrule 
SMVS                                             & HFR( no noise ) [m]    & HFR (noise)  [m]       & injection [m]        & HFR( no noise ) [deg]  & HFR (noise)  [deg]     & injection [deg]      \\ \midrule
-10000 \textless $\mathcal{S}$ \textless -6000 & 0.386±0.221          & 0.282±0.114          & 0.328±0.159          & 2.407±0.932          & 1.807±0.594          & 2.188±1.173          \\
-6000 \textless $\mathcal{S}$ \textless -3000  & 0.482±0.165          & 0.305±0.107          & 0.394±0.017          & 2.738±1.043          & 1.794±0.514          & 1.962±0.501          \\
-3000 \textless $\mathcal{S}$ \textless -1000  & 0.296±0.136          & 0.311±0.160          & 0.346±0.115          & 1.768±0.588          & 2.013±0.875          & 2.054±0.588          \\
-1000 \textless $\mathcal{S}$                  & \textbf{1.848±3.805} & \textbf{3.761±5.985} & \textbf{2.954±5.132} & \textbf{22.38±51.23} & \textbf{22.18±42.82} & \textbf{31.86±58.50} \\ \bottomrule
\end{tabular}
\end{table*}


\noindent \textbf{Results:}
Table \ref{tab_APE} summarizes the relationship between the SMVS and the resulting position deviation. For all localization algorithms and attack methods, the APE remained low when the SMVS was below -1000. However, when the score exceeded -1000, the APE increased dramatically in both translational and rotational directions. This suggests that the SMVS is an effective indicator of SLAM vulnerability.

For the KISS-ICP and hdl-localization, retaining salt-and-pepper noise during HFR attacks yielded greater attack effectiveness. This indicates that large-scale salt-and-pepper noise can confuse localization algorithms, suggesting that incorporating sufficient denoising in preprocessing steps enhance security.
Across all algorithms, injection attacks had higher impacts than removal attacks, revealing that localization systems are more vulnerable to injection attacks than removal attacks. We attribute this to the fact that partial disappearance of geometric features often occur due to occlusion and is covered by SLAM's robustness. In contrast, the injection of structured point clouds is more challenging to mitigate.



\subsection{Evaluation of Spoofer Placement Optimization}

\noindent \textbf{Experimental Settings:}
To evaluate the effectiveness of our spoofer placement optimization method, we conducted simulations comparing three spoofer deployment strategies for vehicles traveling on a course similar to Fig.\ref{header_fig}. The strategies compared were: randomly placing a single spoofer 
200 times within the range of X,Y=90m and 50m 
, targeted placement at the top-10 highest SMVS locations along the trajectory, and using the method proposed in \S \ref{sec:attack_loc_sel}. 


\noindent \textbf{Results:}
As shown in Table \ref{decided_trans}, our proposed method resulted in greater positioning errors compared to the other methods. When compared to the random placement strategy, our method produced 12.6 times larger errors for HFR-based attacks and 15.4 times larger errors for injection attacks. We believe this significant increase in positioning errors is due to our proposed spoofer placement technique, which quantifies vulnerabilities in scan matching using SMVS and determines optimal placement locations to maximize the consequences.\vspace{-0.07in}

\begin{table}[tb]
\centering
\setlength{\tabcolsep}{1.5pt}
\caption{Comparison of estimated position error by spoofer position determination methods (Metrics:APE, Unit:m)}
\label{decided_trans}
\begin{tabular}{c|ccc} 
\toprule
               & Random [m]    & Top-10 SMVS [m] & \textbf{SLAMSpoof (ours)} [m] \\ \midrule
HFR (noise)    & 1.122±3.029         & 0.375±0.190                    & \textbf{14.13±3.684}      \\
Injection      & 0.714±1.759         & 0.401±0.164                    & \textbf{11.00±3.684}       \\ \bottomrule
\end{tabular}
\end{table}

\subsection{Physical-World Evaluation}
\vspace{-0.05in}
Finally, we evaluate the effectivness of the SLAMSpoof on a ground vehicle with LiDAR.

\begin{figure}[tb]
  \centering
  \includegraphics[trim=0mm 0mm 0mm 40mm, clip, scale=0.3]{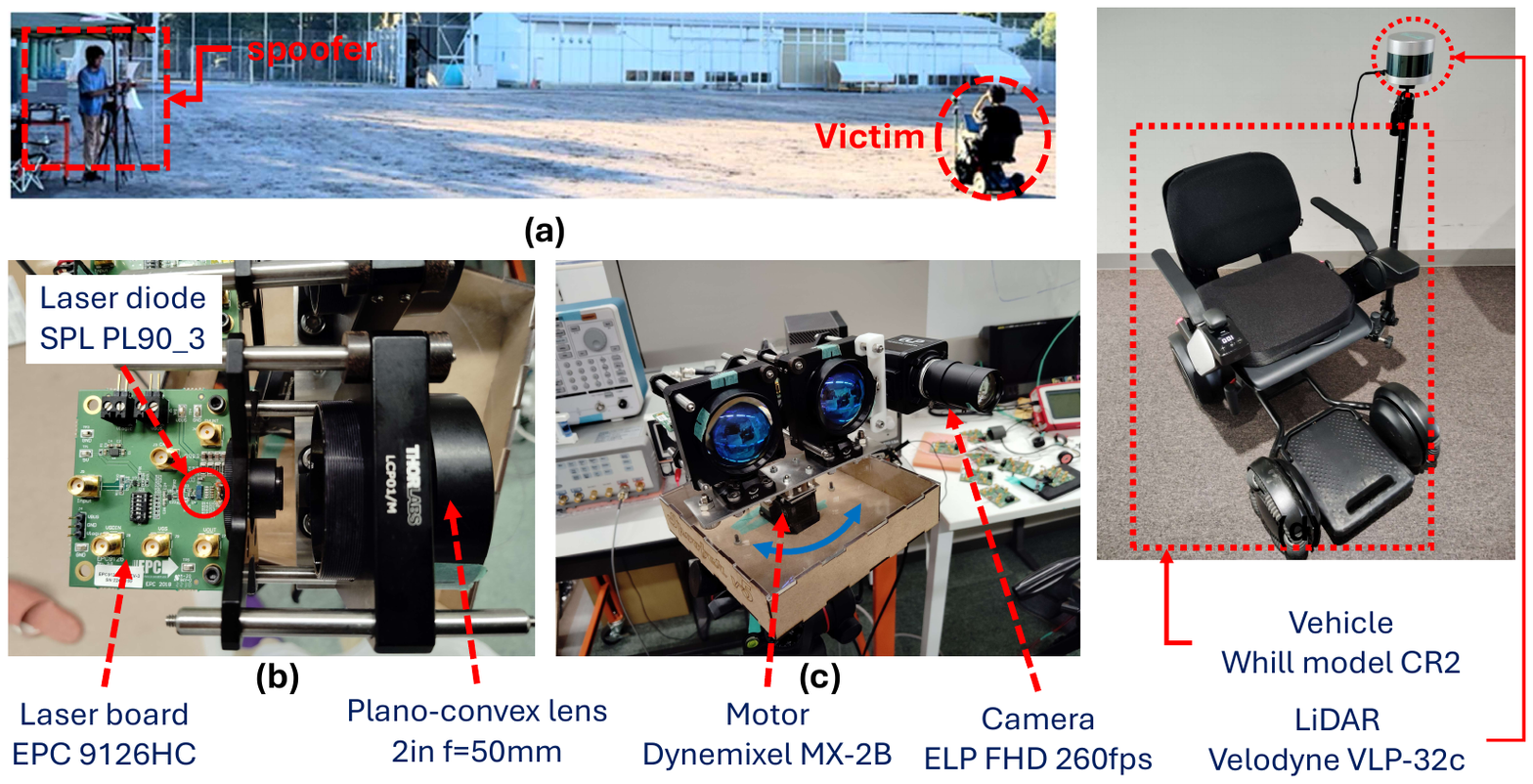}
  \vspace{-0.5in}
  \caption{
  (a) Outdoor experiment setup: The spoofer targets the victim's vehicle $\leq$50 meters away; (b) Spoofer configurations: laser beam with the same wavelength as the target LiDAR is generated by the diode and collimated by the lens; (c) Tracker device~\cite{suzukiwip}: it can detect the victim LiDAR with an IR camera and track it with the rotating table on the servo motor; (d) The victim ground vehicle with LiDAR.
  }
  \label{physical exp}
\end{figure}

\noindent \textbf{Experimental Settings:}
We equipped the target robot with a VLP-32c LiDAR sensor as shown in Fig.\ref{physical exp}. To vary the SMVS within the course, the maximum range was limited to 50 meters. We utilized the attack device proposed in \cite{suzukiwip}, which is capable of pursuing and attacking moving objects outdoors at distances of up to 50
meters. This setup allowed us to test the effectiveness of our method against a mobile target in a realistic outdoor environment.
The attack was conducted as outlined in Sec.III. We employed the same localization algorithm and attack methodology as in our previous simulation experiments.
To evaluate the induced position displacements, we calculated the RMSE between the expected path under attack conditions and the nearest points on the benign path. Due to the varying coordinate systems used by different methods, we converted the predicted trajectories of other techniques to the coordinate system of hdl-localization for a standardized evaluation.



\noindent \textbf{Results:}
As shown in Fig.\ref{header_fig}, our experiments resulted in critical position displacements across all localization methods adopted in this study. Table \ref{e2e_result} presents the quantitative results, demonstrating that regardless of the attack technique or SLAM method employed, we were able to induce estimation errors on the order of several meters. This finding conclusively proves that LiDAR spoofing can significantly impact estimated positions in real-world scenarios.

While simulation results favored injection attacks, real-world experiments proved removal attacks more effective. This discrepancy likely stems from synchronization challenges in the receiving system, limiting the amount of injectable point cloud data. In practice, only 8-10 layers could be injected, despite the target LiDAR performing 32-layer vertical scans.




\begin{table}[tb]
\centering
\caption{Physical-world evaluation results}
\label{e2e_result}
\begin{tabular}{c|ccc} 
\toprule
               & KISS-ICP\cite{vizzo2023ral}    & A-LOAM\cite{zhang2014loam} & hdl\_locatization\cite{koide2019portable} \\ \midrule
HFR     & 8.379 m       & 6.127 m                   & 4.217 m     \\
Injection      & 4.781 m & 4.075 m           & 2.777 m      \\ \bottomrule
\end{tabular}
\end{table}

\section{DISCUSSIONS}
\vspace{-0.05in}
\noindent \textbf{Countermeasures:}
Our study on SLAMSpoof attacks highlights the need for robust defense mechanisms in LiDAR-based localization systems. Potential countermeasures include LiDAR systems with pulse signature technology and sensor fusion with IMUs. Pulse signatures have shown promise against HFR and injection attacks~\cite{sato2024lidar}. Thus, implementing such features in the LiDAR could also provide effective defense against SLAMSpoof. IMU fusion, as implemented in algorithms like LIO-SAM\cite{shan2020lio}, may offer resilience against spoofing by maintaining accurate positional data even when LiDAR is compromised.

\noindent \textbf{Social Impact:} We strongly recommend that the robot developer should not solely rely on LiDAR-based localization in the safety and security critical scenarios although LiDAR-based localization is so far the only choice to achieve centimeter-level accuracy required for autonomous driving. Our SMVS can be used to assess the vulnerability of the operational sites.  Robot developers may consider additional sensors for localization (e.g., IMU and GNSS) If SMVS is so high (e.g., $\geq$-1,000) in their application sites while it may need additional costs.\vspace{-0.1in}


\section{CONCLUSION}
\vspace{-0.05in}
In this study, we developed SLAMSpoof, the first practical LiDAR spoofing attack on localization systems for self-driving vehicles. By manipulating geometric features in point cloud data, our method induces severe localization errors that can force vehicles off-road or bypass critical traffic information. We introduced the Scan Matching Vulnerability Score (SMVS), a novel metric that quantifies vulnerabilities in scan matching algorithms and demonstrates strong correlation with attack outcomes across diverse localization methods.
Our real-world experiments validate the attack effectiveness, resulting in dangerous localization errors of $\geq$4.2 m—exceeding typical lane widths—across three popular LiDAR-based SLAM algorithms. 
Our work provides valuable insights for developing more resilient localization technologies, essential for ensuring the safety and reliability of autonomous vehicles in real-world environments.


\bibliographystyle{IEEEtran}
\bibliography{citation.bib}
\end{document}